\documentclass[10pt,twocolumn,letterpaper]{article}

\usepackage{cvpr}
\usepackage{times}
\usepackage{epsfig}
\usepackage{graphicx}
\usepackage{amsmath}
\usepackage{amssymb}
\usepackage{pifont}

\usepackage{multirow}
\usepackage{paralist}

\usepackage[pagebackref=true,breaklinks=true,letterpaper=true,colorlinks,bookmarks=false]{hyperref}

\cvprfinalcopy 

\def\cvprPaperID{1915} 

\begin{document}

\title{
Generalized Zero-Shot Recognition based on Visually Semantic Embedding}

\author{Pengkai Zhu, Hanxiao Wang, Venkatesh Saligrama\\
Electrical and Computer Engineering Department, Boston University\\
{\tt \small \{zpk, hxw, srv\}@bu.edu}
}

\maketitle

\begin{abstract}
We propose a novel Generalized Zero-Shot learning (GZSL) method that is agnostic to both unseen images and unseen semantic vectors during training. Prior works in this context propose to map high-dimensional visual features to the semantic domain, which we believe contributes to the semantic gap. To bridge the gap, we propose a novel low-dimensional embedding of visual instances that is ``visually semantic.'' Analogous to semantic data that quantifies the existence of an attribute in the presented instance, components of our visual embedding quantifies existence of a prototypical part-type in the presented instance. In parallel, as a thought experiment, we quantify the impact of noisy semantic data by utilizing a novel visual oracle to visually supervise a learner. These factors, namely semantic noise, visual-semantic gap and label noise lead us to propose a new graphical model for inference with pairwise interactions between label, semantic data, and inputs. We tabulate results on a number of benchmark datasets demonstrating significant improvement in accuracy over state-of-art under both semantic and visual supervision.    
\end{abstract}

\section{Introduction}
Zero-shot learning (ZSL) is emerging as an important tool for large-scale classification~\cite{ILSVRCarxiv14}, where one must account for challenges posed by non-uniform and sparse annotated classes \cite{Bhatia15}, the prohibitive expense in labeling large fractions of data \cite{antol2014zero}, and the need to account for appearance of novel objects for in-the-wild scenarios. 

ZSL proposes to learn a model for classifying images for ``unseen'' classes for which no training data is available by leveraging semantic features, which are shared by both seen and unseen classes. A few recent works~\cite{Chao2016AnES,xian2018zero} point out that, unseen image class recognition, while important, overlooks real-world scenarios, where both seen and unseen instances appear. Consequently, generalized zero-shot learning (GZSL) methods capable of recognizing both seen and unseen instances at test time are required.      

We propose to train a GZSL method, that takes labeled seen class images and associated semantic side information as input, while being agnostic to both unseen images and unseen associated semantic vectors.

\noindent {\bf Challenges.} We list challenges in this context:

\noindent \emph{Visual $\rightarrow$ Semantic Gap.} Visual feature representations such as the final-layer outputs of deep neural networks are high-dimensional and not semantically meaningful. This limits the learner in identifying robust associations between visual patterns and semantic data.
    
\noindent \emph{Semantic$\rightarrow$Visual Gap.} A fundamental drawback of semantic data is that they are often not visually meaningful and it is difficult for a learner to identify and suppress non-visual semantic components during training.
Additionally, semantic information provided for some classes (ex. sofa-chair), are nearly identical. This is challenging in a GZSL setting particularly when one of such classes is among the unseen. 

\noindent {\bf Novelty.} At a conceptual level visual representation and supervision fundamentally impacts accuracy. We re-examine these concepts and propose novel methods in Sec. 3 to identify and bridge the visual-semantic gap. 

\noindent \underline{Visually Semantic Embedding.} By a visually semantic embedding, we mean a mapping of visual instances to a representation that mirrors how semantic data is presented for an instance. In Sec. 3.1 we propose to train a model that learns a finite list of parts based on a multi-attention model and expresses the input as a finite probabilistic mixture of part-types, which we then output as our representation. Our intuition is informed by semantic data where for each instance, an annotator could score existence of attributes from a common vocabulary. Analogously, our embedding scores existence of proto-typical part types in a presented instance.

\noindent \underline{3-Node Graphical Model.} 
A key aspect of our setup, which is presented in Sec. 3.1, is a graphical model\cite{Koller:2009:PGM:1795555} that has semantic (S), input (X) and label (Y) variables in a 3-node clique. This is based on the key insight that the labels are not fully explained by either the input or the semantic instance and thus we require a model that accounts for 3-way connection ($S\leftrightarrow X, X\leftrightarrow Y, Y\leftrightarrow S$). This is a significant departure from existing works~\cite{10.1109/TPAMI.2013.140}, where the semantic signal is given paramount importance and the structure is a chain $X\leftrightarrow S \leftrightarrow Y$.  

Following convention, we conduct experiments with semantic supervision on benchmark datasets in Sec. 4. While we demonstrate significant improvement over state-of-the-art, we are driven to understand and quantify how semantic noise can explain GZSL performance loss.

\noindent \underline{Visually Semantic Supervision.}
As a thought experiment we propose to train a novel visual oracle for GZSL supervision. Our intuition is that a visual oracle can reduce semantic noise and provide more definitive feedback about the presence/absence of prototypes. To ensure fair comparison between semantic and visual feedback, we first learn a common vocabulary of protypical parts and part-types unbeknownst to the learner. Like a semantic signal, our visual oracle for each input instance provides the learner only a list of similarity scores, with no other additional description of what the components in the list mean. 

Visual supervision naturally leads us to propose visual evaluation, which involves evaluating predicted visually semantic outputs against the ground-truth. In Sec. 4, we show that GZSL performance improves not only for our method but also for a well-known baseline method~\cite{frome2013devise} when we substitute visual in place of semantic supervision. 

\begin{figure*}[t]
    \centering
    \includegraphics[width=\textwidth]{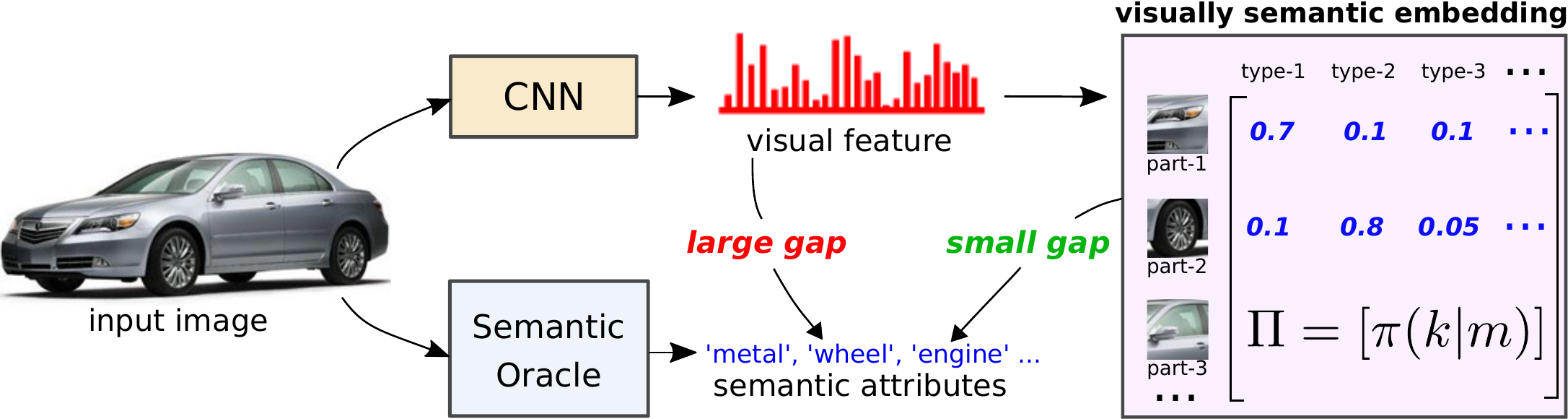}
    \label{fig:sfig1}
    \caption{\small Many existing works attempt to transfer high-dimensional visual features into semantic domain leading to significant visual-semantic gap. To bridge the gap we propose a new latent visual embedding that is {\it visually semantic}. As illustrated our new representation is low dimensional and its components are likelihoods of a part-type relative to proto-typical part-types found across all instances. We posit that our embedding mirrors how semantic components score similarity of an attribute found in an instance in the visual domain.}
    \vspace{-3mm}
\end{figure*}

\section{Related Work}
Zero-shot learning as a topic has evolved rapidly over the last decade and documenting the extensive literature here is not possible. As summarized in \cite{xian2018zero} many existing methods can be grouped into attribute methods as exemplified in \cite{10.1109/TPAMI.2013.140} that leverage attributes as an intermediate feature space to link different classes, embedding methods \cite{frome2013devise} that directly map visual domain to semantic space, and hybrid methods~\cite{zhang2015zero}, that map semantic and visual domain into a shared feature space. Recent work~\cite{Chao2016AnES,xian2018zero} introduces GZSL problem and developed calibration and evaluation protocols showing significant drop in accuracy between ZSL and GZSL. 
Many recent works are beginning to focus attention on the GZSL setup. 

In this context, we propose an embedding based method and describe closely related setups and concepts that have appeared in the literature. We first categorize existing work based on problem setup and different types of side information utilized during training. There are a number of recent works that propose approaches for both ZSL and GZSL cases  ~\cite{Li_2017_CVPR,Annadani_2018_CVPR,zhu2018generative,Verma_2018_CVPR,Xian_2018_CVPR,Jiang_2018_ECCV,chen2018zero,Wang_2018_CVPR,Lee_2018_CVPR}. Among these, there are works that leverage some form of unseen class information during training~\cite{zhu2018generative,Verma_2018_CVPR,Xian_2018_CVPR,Jiang_2018_ECCV} to synthesize unseen examples by means of GAN or VAE training. Other works employ knowledge graphs~\cite{Lee_2018_CVPR,Wang_2018_CVPR} incorporating both seen and unseen classes during training to infer classifiers for unseen classes. Still others are transductive, namely, at test-time \cite{Li_2017_CVPR} they leverage a batch of test examples to further refine their model. While these proposed approaches that leverage unseen class information are interesting, we take the view that for applications involving recognition in-the-wild scenarios, novel classes may only appear at test-time, and it is important to consider such situations. Ultimately, as a subject of future work, it would be interesting to incorporate cases where some unseen class information is known during training, while leaving open the possibility of existence of novel classes at test-time. 

Like us, there are works ~\cite{chen2018zero,Annadani_2018_CVPR} focusing on ZSL and GZSL, while being agnostic to any unseen class information during training. In \cite{Annadani_2018_CVPR}, authors propose an encoder-decoder network with the goal of mirroring learnt semantic relations between different classes in the visual domain. While the goal is similar, our  approach is significantly different. We propose to mirror information provided by semantic attributes visually by means of a low-dimensional statistical embedding. Our embedding scores existence of prototypical part types, where the prototypical part types are learnt from training data. 

In \cite{chen2018zero}, the authors propose an approach that extends methods of \cite{kodirov2017semantic}. Their idea is to penalize approximation error in reconstructing visual domain features from the semantic domain, in addition to penalizing classification loss. Their claim is that by doing so they can overcome semantic information loss suffered in methods that are based on visual to semantic embedding and prevent situations where attributes possibly corresponding to unseen examples maybe lost during training. In contrast, our claim is precisely that many semantic attributes are not visual and conventional visual features are not represented in the presented semantic vectors. Consequently, we propose approaches that on the one hand produces semantically closer visual representations through low-dimensional graphical models, and on the other hand suppress semantic components that are visually unrepresentative by means of discriminative loss functions. In general we do not require high-dimensional estimation, which these methods require.

In this context our approach bears some similarities to \cite{Li_2018_CVPR} and \cite{zhu2018generative}. In particular, \cite{Li_2018_CVPR} propose zoom-net as a means to filter-out redundant visual features such as deleting background and focus attention on important locations of an object. \cite{zhu2018generative} further extend this insight and propose visual part detector (VPDE-Net) and utilize high-dimensional part feature vectors as an input for semantic transfer.  \cite{zhu2018generative}'s proposal is to incorporate the resulting reduced representation as a means to synthesize unseen examples leveraging knowledge of unseen class attributes. Different from these works we develop methods to learn a statistical representation of mixture proportions of latent parts. Apart from being low-dimensional the mixture proportions intuitively capture underlying similarity of a part-type to other such part-types found in other classes. The focus of semantic mapping is then to transfer knowledge between mixture proportion of part types and semantic similarity. 

\begin{figure}[t]
    \centering
    \includegraphics[width=0.35\textwidth]{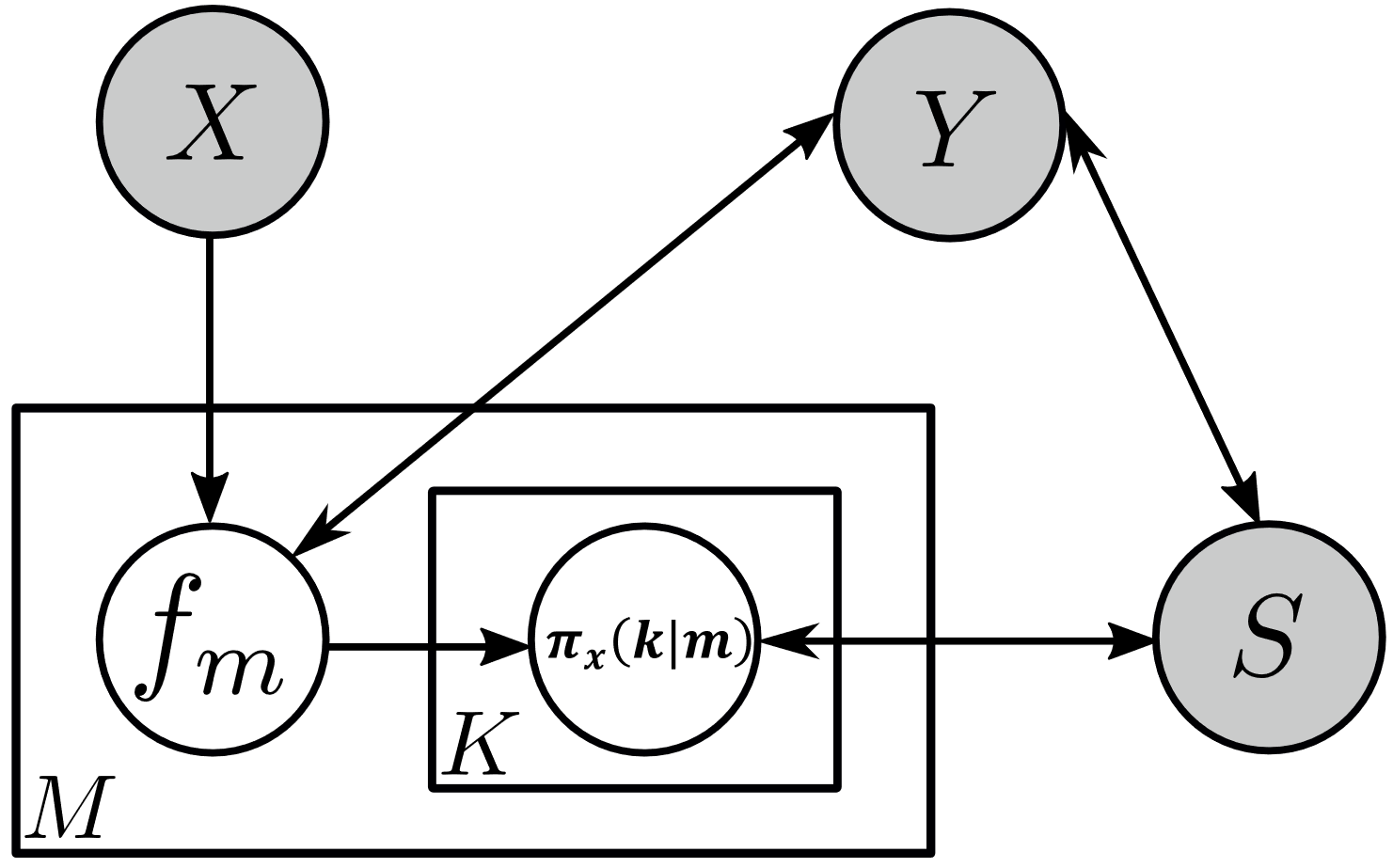}
    \caption{\small Graphical Model of label, semantic signal, and input as a cycle. Input is transformed into a finite collection of feature vector parts indexed by items in a part-list. Feature parts are mapped into structured probability space, with component $\pi_x(k|m)$ denoting the probability of $k$-th type in item $m$ in the part-list. $\Pi_k(x)=[\pi_x(k|m)]$ denoting visual embedding for part $k$.}
    \label{fig:lsm}
\end{figure}

\section{Proposed Approach}
\subsection{A Probabilistic Perspective of GZSL}
\label{sec:PPGZSL}
Let us motivate our approach from a probabilistic modeling perspective. This will in turn provide a basis for our discriminative learning method. $[N]$ denotes integers from $1$ to $N$. Overloading notation we denote $[\eta_{k,m}]$ to mean the matrix as $k,\,m$ range over their values. 
Following convention we denote random variables with upper case letters and a realization by lower case letters. Let $x \in {\cal X}$ be inputs taking values in an arbitrary feature space and ${\cal Y}$ the space of objects or classes. The set ${\cal Y}$ is partitioned into $o \in {\cal O}$ and $u \in {\cal U}$ denoting the collection of observed class labels and unobserved labels respectively.  Associated with each observed and unobserved class labels are semantic signals, $s_o,s_u \in {\cal S}$, taking values in a general space respectively. We denote by $p$ the joint density or marginal densities wherever appropriate. 

Given training data $(x_1,y_1,s_{y_1}),\ldots,(x_n,y_n,s_{y_n}) \subset {\cal X}\times {\cal O}\times {\cal S}$, the task of GZSL is to accurately predict a label with input drawn from $x \stackrel{d}{\sim} p_X(\cdot)$. If we had knowledge of joint probability density, the optimal predictor is the MAP estimate, $y_{\mathrm{MAP}}(x)=\arg\max_{y} \max_{s} \log(p(y,s\ |\ x))$. 
Existing work \cite{10.1109/TPAMI.2013.140} posits instead a chain $X-S-Y$, namely, conditioned on semantic signals, the input and class labels are independent or that the chain $X-Y-S$ \cite{Zhang_2017_CVPR} is true. Nevertheless, we take the view that, {\it since semantic information is not fully visual, visual features are not fully semantic, and labels are not fully captured by either visual or semantic signals}, the three random variables form a cycle in a graphical model. By means of potential functions for graphical models we decompose  
$\log(p(y,s|x) 
\propto \phi_{YX}(y,x)+\phi_{XS}(x,s))+\phi_{SY}(s,y)$.

\noindent {\bf Latent Graphical Model.}
Fig.~\ref{fig:lsm} presents a detailed framework of our model. The input $X=x$ is mapped into a finite set of feature vector parts, $f_m(x) \in \mathbb{R}^C$ indexed by discrete part-list, $m \in [M]$. Feature parts are derived from a multi-attention model with different items $m,m' \in [M], m \neq m'$ focusing on different regions in the image. We then model each feature part vector as a C-dimensional Gaussian Mixture Model, $f_m(x)\approx \sum_i \pi_x(k|m) {\cal N}(\theta_{k,m},\gamma^2 I)$, with isotropic components. 
Note that the parameters, $\theta_{k,m} \in \mathbb{R}^C$, are part and type dependent but shared among all instances. We refer to $\theta_{k,m}$'s as {\it prototypical part-types}. 
We collect the parameters into a matrix $\Theta_m=[\theta_{k,m}]\in \mathbb{R}^{C\times K}$.

Each mixture component, $\pi_x(k|m)$ represents the probability of type $k$ conditioned on part $m$. In this way, the input $x$ is embedded into a collection of mixture components $\Pi_m(x)=[\pi_x(k|m)]$ and $\Pi(x)=[\Pi_m(x)]$. We decompose the likelihood as:
\begin{align} \label{eq.decomp}
\log(p(y,s\mid x)) &\propto \phi_{SX}(s,\Pi(x)) + \phi_{XY}([f_m(x)],y) \\ \nonumber
&+\phi_{YS}(y,s) - \sum_{m=1}^M L_{mix}(\Theta_m,\Pi_m(x), f_m(x))\\ \nonumber
&- L_{prt}([f_m(x)]) 
\end{align}
where, the last two terms respectively model mixture likelihood and enforce diversity of multi-attention of parts. The goal of training algorithm is to estimate the potential functions, $\phi$, the feature part backbone, $f_m(\cdot)$, and the probability maps, $\Pi_m(\cdot)$ by leveraging training data. 

\noindent{\it Latent Visual Embedding Intuition.}
Note that, by design (see Fig.~\ref{fig:lsm} and Eq.~\ref{eq.decomp}), the semantic random vector interacts with the input only through the mixture component $\Pi(x)$. This is in contrast to existing works where the interaction is high-dimensional. Our intuition is that, just like a component of a semantic vector quantifies existence of an attribute, in an analogous fashion, $\pi_x(k|m)$ quantifies existence of part-type $k$ in part $m$. While prototypical parts such as $\theta_{k,m}$ are high-dimensional, the corresponding mixture component $\pi_x(k|m)$ is a scalar number. In this way we propose to reduce the semantic-visual gap by removing irrelevant visual features that are not transferable.

\noindent {\bf Visual Oracle Supervision (VOS).}
We consider visual oracles capable of providing feedback for learner predicted visual embedding. We denote $\Pi_{vo}(x)$ as oracle feedback. As to how to build a visual oracle will be discussed later. We can consider structured and class-averaged VOS. In the structured version, for each instance, $x$, VOS reveals the probabilistic embedding $\Pi_{vo}(x)$; 
and in the class-averaged case, only reveals  $\bar{\Pi}_{vo}=\mathbb{E}_{X\mid Y}[\Pi_{vo}(x) \mid y]$. The main difference in Eq.~\ref{eq.decomp} is that we substitute oracle parameters for the semantic signal, i.e., $s$ with $\Pi_{vo}(x),\, \bar \Pi_{vo}$ etc.  

\noindent{\it Justification of Visual Oracle Supervision.}
In constructing the visual oracle, our goal is driven by the need to quantify semantic noise., To do so we need an oracle that provides no more information than a ``noiseless'' semantic one. This is not that hard since our visual oracle presents the learner with mixture values $\Pi_{vo}(x)$ with out identifying what these numbers mean or which classes, parts or locations they refer to. All that a learner knows is that what the oracle is communicating information that has definitive visual meaning.

\noindent{\bf Visual and Semantic Test-Time Evaluation.}
At test-time, following convention, for the semantic setting, we assume that the codebook consisting of seen and unseen semantic attribute vectors, $\{s_y| y\in {\cal Y}\}=\{s_o| o \in {\cal O}\} \cup \{s_u| u \in {\cal U}\}$ are revealed to the learner. For a test image, $x$, the learner must identify the hidden label. To do this, the learner computes $\Pi(x)$, and estimates the label by maximizing the visual-semantic potential $\hat y(x)=\arg\max_{y\in {\cal Y}}\phi_{SX}(s_y,\Pi(x))$. In the visual evaluation setting, the class-level $\{\bar{\Pi}_{vo}^{y}|y\in \mathcal{Y}\}$ is revealed during test time and the learner make prediction by maximizing the visual-semantic potential in the $\Pi$ space $\hat{y}(x) = \arg\max_{y\in\mathcal{Y}} \phi_{SX}(\bar{\Pi}_{vo}^y, \Pi(x)).$

\subsection{Model and Loss Parameterization}
\label{sec:param}
\noindent \textbf{Part Feature model $f_m(\cdot)$}: Inspired by \cite{zheng2017learning}, we use a multi-attention convolutional neural network (MA-CNN) to map input images into a finite set of feature vector parts, $f_m(x)$. Specifically, $f_m(x)=[f_{m,c}]$ contains a feature extractor $E$ and a channel grouping model $G$, where $E(x) \in \mathbb{R}^{W \times H \times C}$ is a global feature map, and $G(E(x)) \in \mathbb{R}^{M\times C}$ is a channel grouping weight matrix. We then calculate an attention map $A_m(x) \in \mathbb{R}^{W \times H}$ for the $m$-th part:
\begin{equation}
    A_m(x) = \mathrm{sigmoid}\big(\sum_c G_{m,c}(x) \times E_c(x)\big)
\end{equation}
The part feature $f_m(x)\in \mathbb{R}^C$ is then calculated as:
\begin{equation}
f_{m,c} (x) = \sum_{w,h} [A_m(x) \odot E_c(x)]_{(w, h)}  , \quad \forall c \in [C]
\end{equation}
where $\odot$ is the element-wise multiplication. We parameterized $E(\cdot)$ by the ResNet-34 backbone (to $conv5\_x$), and $G(\cdot)$ by a fully-connected layer.

\noindent \textbf{Mixture model $\Pi(\cdot)$}: Note that our Gaussian mixture model implies:
$$
\mathbb{E}_Z[f_m(x)| \Pi_m, \Theta_m] = \Theta_m \Pi_m(x)
$$
This sets up a matrix factorization problem, with positivity constraints on the components of $\Pi_m(x)$. Observe that, at test-time, since the matrices $\Theta_m$ are known, the solution to $\Pi(x)$ reduces to solving a linear system of equations with positivity constraints. Alternatively, we can employ a Bayesian perspective (which is what we do) and compute: 
\begin{equation} \label{eq:Pi}
    \pi_x(k|m) \propto \widehat{\pi(k|m)} {\cal N}(f_m(x);\theta_{k,m},\gamma^2I)
\end{equation}
where, $\widehat{\pi(k|m)}$ is the prior for prototype $k$ in part $m$ estimated during training. 

\noindent \textbf{Part Feature Learning Loss $L_{prt}$}: To encourage a part-based representation $f_m(x)$ to be learned, we follow \cite{zheng2017learning}.  Since $f_m(x)$ can be decomposed into $A_m(x)\odot E(x)$, we want to force the learned attention maps $A_m$ to be both  compact within the same part, and divergent among different parts. We define $L_{prt} ([f_m(x)])$ to be:
\begin{equation}
    L_{prt} ([f_m(x)]) = \sum_m (L_{dis} (A_m (x)) + \lambda L_{div} (A_m (x))) \label{eq:loss_prt}
\end{equation}
where the compact loss $L_{dis} (A_m)$ and divergent loss $L_{div} (A_m)$ are defined as ($x$ is dropped for simplicity):
\begin{align}
     L_{dis}({A_m}) &= \sum_{w,h} A_{m}^{w,h}[\|w - w^*\|^2 + \|h - h^*\|^2] \label{eq:loss_Dis} \\
     L_{div}({A_m}) &= \sum_{w,h} A_{m}^{w,h}[max_{n, n\ne m}A_n^{w,h} - \zeta] \label{eq:loss_Div}
\end{align}
where $A_{m}^{w,h}$ is the amplitude of $A_m$ at coordinate $(w,h)$, and $(w^*, h^*)$ is the coordinate of the peak value of $A_m$, $\zeta$ is a small margin to ensure the training robustness.

\noindent \textbf{Mixture Model Learning Loss $L_{mix}$}: We pose this as a standard max-likelihood estimation problem, and learn $\theta_{k,m},\gamma$ parameters using the EM algorithm to fit the feature vectors $f_{m,i}\triangleq f_m(x_i)$, with $x_i$ being training examples. We can write 
%
the negative log-likelihood for i-th sample:
\begin{equation}
\resizebox{.45 \textwidth}{!} 
{$
L_{mix}(\Theta_m, \Pi_m,f_{m,i})  = - \log ( \sum_k \bar{\pi}(k|m) {\cal N}(f_{m,i}; \Theta_m, \gamma^2 I))$}
\end{equation}
where the parameters are optimized by the Expectation-Maximization (EM) algorithm during training. Once they are learned, the mixture component embedding $\Pi(x)$ can be inferred with Eq.(\ref{eq:Pi}).

\noindent \textbf{Semantic-Label Potential $\phi_{YS}$}: In the GZSL problem, we usually assume a deterministic one-to-one mapping between the semantic signals to class labels provided by a semantic oracle (human annotator). The Semantic-Label potential function is thus simply modeled by an indicator function:
\begin{equation}
    \phi_{YS}(y, s_{y'}) = \mathbb{I} (y = y') 
\end{equation}

\noindent \textbf{Visual-Label Potential $\phi_{XY}$}: To map visual representations to class labels, we construct a classification model $D$ that takes the concatenated part features $[f_m(x)]$ as input and outputs a classification prediction, i.e. $D([f_m(x)]) \in \mathbb{R}^{|\mathcal{O}|}$, where $|\mathcal{O}|$ refers to the number of observed classes. In our implementation, $D(\cdot)$ is simply a fully-connected layer followed by a softmax. Let $\hat{p}(y)$ denote the one-hot encoding of the ground-truth class label $y$
for input image $x$, the potential $\phi_{XY}$ is given by the negative cross-entropy between label and prediction:
\begin{equation}
    \phi_{XY} ([fm(x)], y) = - \mathrm{CE} (D([f_m(x)]), \hat{p}(y))
\end{equation}

\noindent \textbf{Visual-Semantic Potential $\phi_{SX}$}: Existing works often take the raw feature vectors as the visual representation, i.e. $E(x)$ or $f_m(x)$, and suffer from a large discrepancy between the visual and semantic domains. To mitigate such a gap, we propose to adopt the latent mixture component embedding $\Pi(x)$. 

\noindent {\it Semantic Oracle.} In the common GZSL setting where the semantic signals are obtained from a human annotator, we construct a Semantic Mapping model $V(\Pi(x))$ to project $\Pi(x)$ into $\mathcal{S}$, where $V(\cdot)$ is further parameterized by a neural network. Given an imput image $x$ and its semantic attribute $s_y$, the potential $\phi_{SX}$ is modeled as:
\begin{align}
\resizebox{.5 \textwidth}{!} 
{$
\phi_{SX} (s_y, \Pi(x)) = - \sum_{y'\in\mathcal{O}} [\eta\mathbb{I}(y'=y) +s_{y'}^\top V(\Pi(x)) 
-s_y^\top V(\Pi(x))]_+ $}
 \label{eq:pi_to_s}
\end{align}
where $\eta$ is a margin parameter.

\noindent {\it Visual Oracle.} As discussed by Sec.\ref{sec:PPGZSL}, we want to evaluate the efficacy of the proposed latent embedding. We thus considers a visual oracle which directly provides $\Pi_{vo}(x)$ as a visually semantic supervision, which is a list of visual part similarity scores. In this case, we no longer need $V(\cdot)$ since both $\Pi(x)$ and $\Pi_{vo}(x)$ are already in the same space. The potential $\phi_{SX}$ is thus:
\begin{equation}
    \phi_{SX} (\Pi_{vo}(x), \Pi(x)) = - |\Pi_{vo}(x) - \Pi(x)|_F^2,
\end{equation}
where $|\cdot|_F$ is the Frobenius norm.

\subsection{Implementation Details}

Our model takes input image size as [448 $\times$ 448] and and the output of $E(x)$ is in the size of 14 $\times$ 14 $\times$ 512. There are 4 parts in the model and in each part, the number of types $M$ is set to 16. $\lambda$ in Eq.(\ref{eq:loss_prt}) and $\zeta$ in Eq.(\ref{eq:loss_Div}) is empirically set to 5 and 0.02. In the semantic-oracle scenario, the smenatic mapping model $V(\cdot)$ is implemented by a two fc-layer neural network with ReLU activation.

In the visual-oracle scenario, the visual oracle is built to provide $\Pi_{vo}(x)$. It consists of the part feature model $f_m(\cdot)$ and a classifier $D$, where the feature extractor $E(\cdot)$ in $f_m(\cdot)$ is parameterized by the VGG-19 convolutional layers. We choose VGG instead of ResNet backbone to avoid that the learner learns to recover the same parameters in the oracle via the $\Pi_{vo}(x)$. The oracle is first trained by maximizing $\phi_{XY} - L_{prt}([f_m(x)])$ to learn a discriminating $f_m(\cdot)$. Then the EM optimization over $\sum_{m=1}^ML_{mix}(\Theta_m \Pi_m, f_m(x))$ is done to generate $\Pi_{ov}(x)$ for our model. During training, the oracle provide instance level $\Pi_{vo}(x)$ and in the test time, only class-averaged $\bar{\Pi}_{vo}$ is revealed to the learner to make the prediction.

\begin{table*}[t]
    \centering
\renewcommand{\arraystretch}{1}
\setlength{\tabcolsep}{0.4cm}
    \begin{tabular}{|l| c c c|c c c|c c c|}
        \hline
        \multirow{2}{*}{\bf{Methods}} & \multicolumn{3}{c|}{\bf{CUB}} & \multicolumn{3}{c|}{\bf{AWA2}} & \multicolumn{3}{c|}{\bf{aPY}} \\
         & ts & tr & H & ts & tr & H & ts & tr & H\\
        \hline
        SJE\cite{akata2015evaluation} & 23.5 & 59.2 & 33.6 & 8.0 & 73.9 & 14.4 & 3.7 & 55.7 & 6.9 \\
        SAE\cite{kodirov2017semantic} & 7.8 & 54.0 & 13.6 & 1.1 & 82.2 & 2.2 & 0.4 & 80.9 & 0.9 \\
        SSE\cite{zhang2015zero} & 8.5 & 46.9 & 14.4 & 8.1 & 82.5 & 14.8 & 0.2 & 78.9 & 0.4 \\
        GFZSL\cite{verma2017simple} & 0.0 & 45.7 & 0.0 & 2.5 & 80.1 & 4.8 & 0.0 & {\bf\color{blue}83.3} & 0.0 \\
        CONSE\cite{norouzi2013zero} & 1.6 & \bf\color{blue}72.2 & 3.1 & 0.5 & 90.6 & 1.0 & 0.0 & \bf\color{red}{91.2} & 0.0 \\
        ALE\cite{akata2016label} & 23.7 & 62.8 & 34.4 & 14.0 & 81.8 & 23.9 & 4.6 & 73.7 & 8.7 \\
        SYNC\cite{changpinyo2016synthesized} & 11.5 & 70.9 & 19.8 & 10.0 & 90.5 & 18.0 & 7.4 & 66.3 & 13.3 \\
        DEVISE\cite{frome2013devise} & 23.8 & 53.0 & 32.8 & 17.1 & 74.7 & 27.8 & 4.9 & 76.9 & 9.2 \\
        PSRZSL\cite{Annadani_2018_CVPR} & 24.6 & 54.3 & 33.9 & 20.7 & 73.8 & 32.3 & 13.5 & 51.4 & 21.4 \\
        SP-AEN\cite{chen2018zero} & \bf\color{blue}34.7 & 70.6 & 46.6 & 23.3 & \bf\color{blue}90.9 & 37.1 & 13.7 & 63.4 & 22.6 \\
        \hline
        Ours(S) & 33.4 & \bf\color{red}{87.5} & \bf\color{blue}48.4 & \bf\color{blue}41.6 & \bf\color{red}{91.3} & \bf\color{blue}57.2 & \bf\color{blue}24.5 & 72.0 & \bf\color{blue}36.6 \\
        Ours($\Pi$) & \bf\color{red} 39.5 & 68.9 & \bf\color{red}{50.2} & \bf\color{red}{45.6} & 88.7 & \bf\color{red}{60.2} & \bf\color{red}{43.6} & 78.7 & \bf\color{red}{56.2} \\
        \hline
    \end{tabular}
    \caption{gZSL learning results on CUB, AWA2 and aPY. ts = test classes (unseen classes), tr = train classes (seen classes), H = harmonical mean. The accuracy is class-average Top-1 in \%.  The highest accuracy is in \textcolor{red}{red} color and the second is in \textcolor{blue}{blue} (better viewed in color).}
    \vspace{-2mm}
    \label{tab:gzsl}
\end{table*}

\section{Experiments}
\noindent \textbf{Datasets.}
We evaluate the performance of our model on three commonly used benchmark datasets for GZSL:  {\em Caltech-UCSD Birds-200-2011} (CUB)~\cite{WahCUB_200_2011}, {\em Animals with Attributes 2} (AWA2)~\cite{xian2018zero} and {\em Attribute Pascal and Yahoo} (aPY)~\cite{farhadi2009describing}. CUB is a fine-grained dataset which contains 200 different types of birds. CUB has 11,788 images and 312-dim annotated semantic attributes. AWA2 is a coarse-grained dataset which has 37,322 images from 50 different animals. 85 binary and continuous class attributes are provided. aPY is also a coarse-grained dataset with 64 semantic attributes. It has 15,339 images of 20 Pascal classes and 12 Yahoo classes. We did not choose SUN~\cite{Xiao:2016:SDE:2963034.2963064} dataset for the reason that the scene images in SUN cannot be easily decomposed into visual parts which are compact and consistent across different scenes, and consequently not expected to benefit from our formulation. The statistics of the datasets are summarized in Table~\ref{tab:data_summary}.

\noindent \textbf{Setting.}
We evaluate performance for both GZSL and ZSL settings. Following the protocol in \cite{xian2018zero}, in the GZSL setting, the average-class Top-1 accuracy on unseen classes (ts), seen classes (tr) and the harmonic mean (H) of ts and tr are evaluated; In the ZSL setting, we report the average-class Top-1 accuracy on both Standard Split (SS) and Proposed Split (PS). 

We examine impact of different concepts such as visual representation, semantic vs. visual supervision on GZSL performance. As summarized by Table \ref{tab:compare_methods}, two variants of the proposed model are evaluated: 
(1) \textbf{Ours(S)}: We benchmark performance of proposed latent visual embedding in the conventional GZSL setting. That is, we train a mapping to project the learned latent representation $\Pi$ into the semantic space $\mathcal{S}$ (Eq.\ref{eq:pi_to_s}) under the supervision provided by the semantic oracle, and during test time the semantic attributes for unobserved classes $\{s_o|o \in \mathcal{O}\}$ are also revealed. {\color{black} For this model, solely semantic supervision is leveraged, as same as all the competing methods.}
(2) \textbf{Ours($\Pi$)}: We quantify the drawbacks of semantic information for GZSL by replacing semantic signals with visual signals $\Pi_{vo}$ generated by our visual oracle, which is then used for both supervision and evaluation.

\begin{table}[t]
    \centering
    \begin{tabular}{l c c c c c}
        \hline
        \bf{Dataset} & \bf{Num Att} & $\mathcal{Y}$ & $\mathcal{O}$ & $\mathcal{U}$ & \bf{Image}\\
        \hline
        CUB\cite{WahCUB_200_2011} & 312 & 200 & 150 & 50 & 11788 \\
        AWA2\cite{xian2018zero} & 85 & 50 & 40 & 10 & 37322 \\
        aPY\cite{farhadi2009describing} & 64 & 32 & 20 & 12 & 15339 \\
        \hline
    \end{tabular}
    \caption{\small Statistics for CUB\cite{WahCUB_200_2011}, AWA2\cite{xian2018zero} and aPY\cite{farhadi2009describing}. Number of semantic attributes, number of class for all($\mathcal{Y}$), seen($\mathcal{O}$) and unseen($\mathcal{U}$), and the number of images are listed.}
    \label{tab:data_summary}
\end{table}

\noindent \textbf{Training Details.}
To train our models, we take an alternative optimization approach where in each epoch, we update the weights in two steps. In step 1, only the weights of $G(\cdot)$ is updated by minimizing $L_{prt}$. In step 2, we freeze the weights of $G(\cdot)$ and update all the other modules. The semantic model (Ours(S)) and the visual oracle is trained by $\phi_{XY}$ while Ours($\Pi$) model is trained by $\phi_{SX} + \phi_{XY}$ in step 2. Adam optimizer is used to optimize the loss in each step. The learning rate for step 1 and step 2 is set to 1e-6 and 1e-5, respectively.

Our models are trained for 80, 60 and 70 epochs on CUB, AWA2 and aPY, respectively. As for the visual oracle, it is trained to 70 epochs on AWA2 and 60 epochs on CUB and aPY. The feature extractor $E(\cdot)$ is initialized with ImageNet pretrained weights. The learning rate for $V(\cdot)$ and $\eta$ in Eq.(\ref{eq:pi_to_s}) is selected via cross-validation. {\color{black} For the optimization of $L_{mix}$, the EM algorithm is terminated if the loss did not change or after 300 steps.}

\begin{table}[t]
    \centering
    \begin{tabular}{l c c c}
        \hline
        \bf{Method} & \bf{Supervision} & \bf{Evaluation} & \bf{Representation}\\
        \hline
        Others & S$_O$ & $\mathcal{S}$ & $\mathcal{F}$ \\
        Ours(S) & S$_O$ & $\mathcal{S}$ & $\Pi$ \\
        Ours($\Pi$) & $\Pi_{vo}$ & $\Pi$ & $\Pi$ \\
        \hline
    \end{tabular}
    \caption{Comparison of the supervision, evaluation embedding and feature representations for our model and others. $\mathcal{S}$: semantic embedding; $\Pi$: latent visual part similarity embedding; $\mathcal{F}$: raw visual feature embedding;  S$_O$: semantic oracle; $\Pi_{vo}$: visual oracle.}
    \label{tab:compare_methods}
\end{table}

\noindent \textbf{Competing Methods.}
To validate the benefits of the proposed latent visual embedding, we compare against other state-of-the-art methods which also utilize semantic supervision and visual representation. Ten competitors are compared: SJE\cite{akata2015evaluation}, ALE\cite{akata2016label}, and DEVISE\cite{frome2013devise} which use structured loss to learn a linear compatibility between visual and semantic space; SSE\cite{zhang2015zero} learns the compatibility function in a latent common space for visual and semantic embedding; GFZSL\cite{verma2017simple} models the the class-conditional distribution as multi-variate Gaussian; CONSE\cite{norouzi2013zero} and SYNC\cite{changpinyo2016synthesized} learns maps the unseen image into semantic representation via combination of seen classes or phantom classes; SAE\cite{kodirov2017semantic} learns the mapping from semantic to visual embedding; PSRZSL\cite{Annadani_2018_CVPR} and SP-AEN\cite{chen2018zero} try to preserve the semantic relations in the mapping by encoder-decoder network or adversarial training.

\begin{table}[tb!]
    \centering
    \begin{tabular}{| l | c c|c c|c c|}
        \hline
        \multirow{2}{*}{\bf{Methods}}& \multicolumn{2}{c|}{\bf{CUB}} & \multicolumn{2}{c|}{\bf{AWA2}} & \multicolumn{2}{c|}{aPY}\\
         & SS & PS & SS & PS & SS & PS \\
        \hline
        SJE\cite{akata2015evaluation} & 55.3 & 53.9 & 69.5 & 61.9 & 32.0 & 32.9\\
        SAE\cite{kodirov2017semantic} & 33.4 & 33.3 & 80.7 & 54.1 & 8.3 & 8.3 \\
        SSE\cite{zhang2015zero} & 43.7 & 43.9 & 67.5 & 61.0 & 31.1 & 34.0 \\
        GFZSL\cite{verma2017simple} & 53.0 & 49.3 & 79.3 & 63.8 & 51.3 & 38.4 \\
        CONSE\cite{norouzi2013zero} & 36.7 & 34.3 & 67.9 & 44.5 & 25.9 & 26.9 \\
        ALE\cite{akata2016label} & 53.2 & 54.9 & 80.3 & 62.5 & 30.9 & 39.7 \\
        SYNC\cite{changpinyo2016synthesized} & 54.1 & 55.6 & 71.2 & 46.6 & 39.7 & 23.9 \\
        DEVISE\cite{frome2013devise} & 53.2 & 52.0 & 68.6 & 59.7 & 35.4 & 39.8 \\
        PSRZSL\cite{Annadani_2018_CVPR} & - & 56.0 & - & 63.8 & - & 38.4 \\
        SP-AEN\cite{chen2018zero} & - & 55.4 & - & 58.5 & - & 24.1 \\ 
        \hline
        Ours(S) & \bf\color{blue} 63.7 & \bf\color{blue} 66.7 & \bf\color{blue} 90.7 & \bf\color{blue} 69.1 & \bf\color{blue} 52.1 & \bf\color{blue} 50.1 \\ 
        Ours($\Pi$) & \bf\color{red}{68.8} & \bf\color{red}{71.9} & \bf\color{red}{92.4} & \bf\color{red}{84.4} & \bf\color{red}{54.4} & \bf\color{red}{65.4}\\
        
        \hline
    \end{tabular}
    \caption{Zero shot learning results on CUB, AWA2 and aPY. SS = standard split, PS = proposed split. The results are class-average Top-1 accuracy in \%. The highest accuracy is in \textcolor{red}{red} color and the second is in \textcolor{blue}{blue} (better viewed in color).}
    \label{tab:zsl}
\end{table}

\subsection{Generalized Zero Shot Learning Evaluation}
The results for the GZSL setting are shown in Table.~\ref{tab:gzsl}. Observe that 
the proposed methods, Ours(S) and Ours($\Pi$) consistently outperforms state-of-the-art methods in the GZSL setting. Specifically, the harmonic mean of the accuracy for seen (tr) and unseen (ts) classes with Ours(S) and Ours($\Pi$) reaches $48.4\%$, $50.2\%$ on CUB, $57.2\%$, $60.2\%$ on AWA2, and $36.6\%$, $56.2\%$ on aPY, which dominate other competing methods and often surpass the third-best result by a very large margin, e.g. a $>20\%$ improvement on AWA2, and a $>10\%$ improvement on aPY. While several competing methods (e.g. \cite{verma2017simple,kodirov2017semantic,norouzi2013zero,zhang2015zero}) only perform well on the seen classes and obtain close-to-zero accuracy on unseen classes, we are able to classify both seen and unseen improving upon existing works in the GZSL setting.

\noindent {\it State-of-art comparison with Semantic Supervision.} Note that under identical conditions of semantic supervision, the gain in our method (ours(S)) can be attributed primarily to our latent visual embedding ($\Pi(\cdot)$). Different from the conventional visual representation, which is a high-dimensional deep CNN feature vector, and not semantically meaningful, $\Pi(\cdot)$ intrinsically describes the input image by a common vocabulary of prototypical parts. These prototypical parts are estimated by the latent mixture model with training images and the components of $\Pi$ quantify the existence of a prototypical part in the instance. Such a representation resembles the semantic similarity of semantic vectors and leads to mitigating the visual-semantic gap. 

\noindent {\it Semantic vs. Visual Supervision.}
Observe that Ours($\Pi$) consistently achieves better performance than Ours(S), e.g. $1.8\%$, $3.0\%$ and $19.6\%$ absolute improvement in the harmonic mean on the three datasets. This comparison shows that, although the proposed latent visual embedding $\Pi$ is able to reduce the Visual$\rightarrow$Semantic gap, the semantic attributes are noisy in that they contain information that are difficult to transfer from the visual domain (e.g. 'smelly', 'agility', 'weak'). Using semantic supervision and evaluation for GZSL thus fundamentally limits attaining high accuracy. By switching to visual supervision and evaluation provided by our visual oracle, we see the potential to further improve GZSL accuracy. This comparison is fair since our visual oracle provides only a list of similarity scores without any other identifying high-dimensional features to the learner. This is similar to the case of a semantic oracle providing attribute annotations.

\noindent {\it Issue with aPY.}
Finally, observe that on aPY most existing methods fail to recognize unseen classes achieving nearly zero accuracy, while we get a significant improvement from Ours(S) and Ours($\Pi$). The reason is that aPY attributes are extremely noisy and are not visually representative  (e.g. bus-car attributes nearly identical). Thus, the semantic supervision cannot provide useful information for training a GZSL model.

\begin{table*}[t]
    \centering
\renewcommand{\arraystretch}{1}
\setlength{\tabcolsep}{2mm}
    \begin{tabular}{|l | c c c | c c |c c c | c c |c c c | c c|}
        \hline
        \multirow{2}{*}{\bf Methods} & \multicolumn{5}{c|}{CUB} & \multicolumn{5}{c|}{AWA2} & \multicolumn{5}{c|}{aPY} \\
         \cline{2-16}
         & ts & tr & H & SS & PS & ts & tr & H & SS & PS & ts & tr & H & SS & PS\\
        \hline
        Ours($\Pi_{flat}$) & 38.4 & \bf 69.8 & 49.6 & 66.8 & 69.5 & 42.6 & \bf 88.7 & 57.6 & 91.7 & 84.0 & 36.5 & \bf 88.7 & 51.7 & 53.6 & 62.9\\
        Ours($\Pi$) & \bf{39.5} & 68.9 & \bf{50.2} & \bf 68.8 & \bf 71.9 & \bf{45.6} & \bf 88.7 & \bf{60.2} & \bf 92.4 & \bf 94.4 & \bf{43.6} & 78.7 & \bf{56.2} & \bf 54.4 & \bf 65.4\\
        \hline
    \end{tabular}
    \caption{\small Test accuracy for different visual representations on CUB, AWA2 and aPY. Ours($\Pi_{flat}$): our model with flatten visual representation. Ours($\Pi$): our model with structured visual representation. The accuracy class-average Top-1 in \%. Both gZSL (ts, tr, H) and ZSL (SS, PS) performances are reported.}
    \label{tab:flat}
    \vspace{-3mm}
\end{table*}

\begin{table}[t]
    \centering
    \setlength{\tabcolsep}{1.5mm}
    \begin{tabular}{|l|c|c|c|c|c|c|}
        \hline
        \multirow{2}{*}{\bf Methods} & \multicolumn{3}{c|}{CUB} & \multicolumn{3}{c|}{AWA2} \\
        \cline{2-7}
         & ts & tr & H & ts & tr & H \\
        \hline
        DEVISE (S) & 23.8 & 53.0 & 32.8 & 17.1 & 74.7 & 27.8 \\
        DEVISE ($\Pi_{vo}$) & \bf{24.6} & \bf{53.3} & \bf{33.7} & \bf{28.3} & \bf{75.3} & \bf{41.2} \\
        \hline
    \end{tabular}
    \caption{gZSL performances for DEVISE\cite{frome2013devise} using semantic supervision and visual oracle supervision ($\Pi_{vo}$).}
    \label{tab:our_attr}
    \vspace{-2mm}
\end{table}

\begin{figure}[t]
    \centering
    \includegraphics[width=\linewidth]{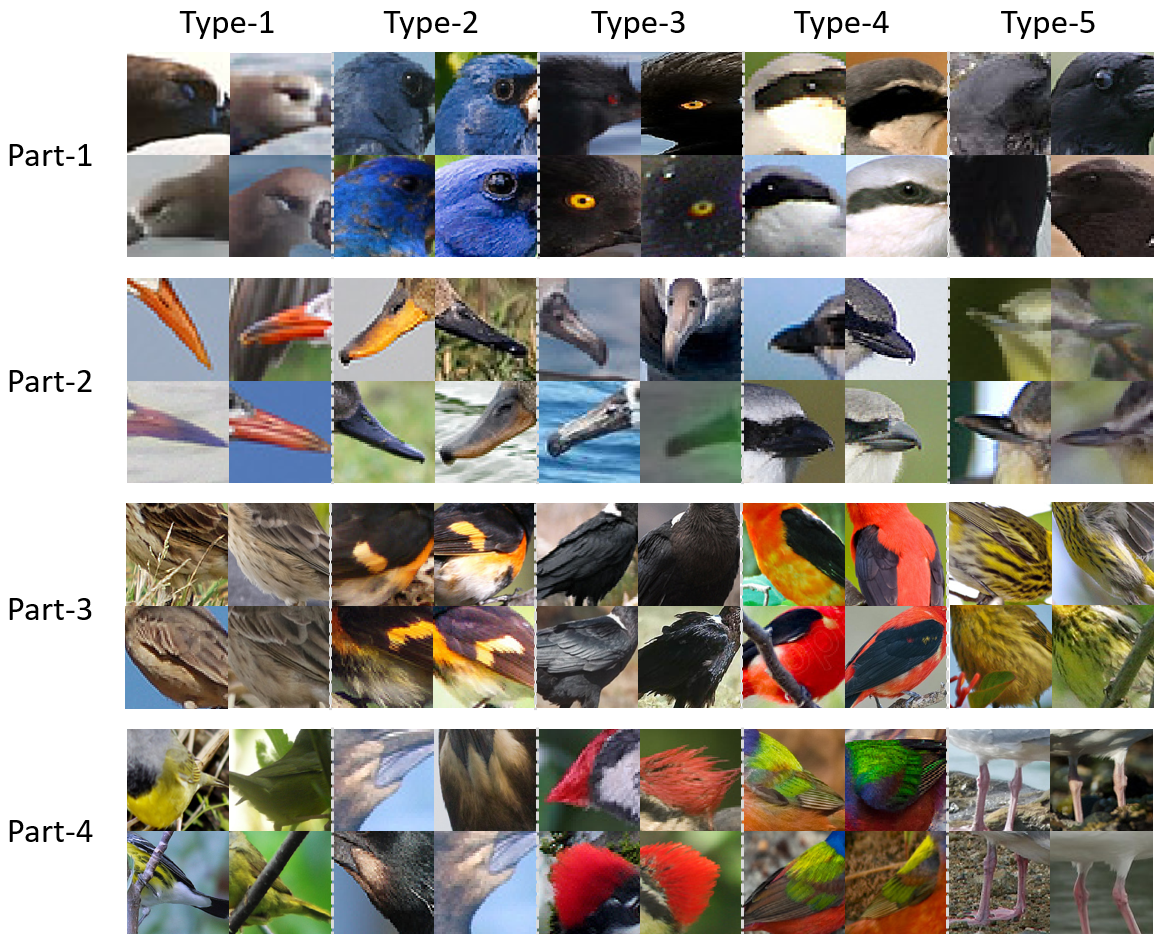}
    \caption{\small Example of types in each part on CUB dataset. Each two rows belong to a part and each two columns belong to a type. In $m$-th part, the example can be labeled with a scalar $\pi(k|m)$ indicating its probability of belonging to the $k$-th type. The example in the type has the largest $\pi(k|m)$ among all types. Note that these types are semantically meaningful and visually distinguishable.}
    \label{fig:cluster_example}
\end{figure}

\vspace{-1mm}
\subsection{Zero Shot Learning Evaluation}
\vspace{-1mm}
We next evaluate the results for the traditional ZSL setting, where only unseen classes are included during testing. The results are reported in Table \ref{tab:zsl}. Observe that many competing methods, which are ineffective in the GZSL setting (e.g. \cite{verma2017simple,kodirov2017semantic,norouzi2013zero,zhang2015zero}), realize a huge performance gain in the ZSL setting. However, Ours(S) and Ours($\Pi$) is robust and still outperforms the competing methods. Again, this can be attributed to our proposed latent visual embedding. Ours(S) model consistently obtains superior performance,  improving the state-of-the-art accuracy from 55.3\% to 63.7\% for SS on CUB, from 80.7\% to 90.7\% for SS on AWA2, and from 51.3\% to 52.1\% for SS on aPY. 
A similar improvement can also be observed on the PS split. This observation shows that, with the same level of semantic supervision, our low-dimensional latent visual embedding is more semantically meaningful than the traditional high-dimensional visual features, and thus effectively bridges the Visual$\rightarrow$Semantic gap. Additionally, with the visual supervision provided by our visual oracle, Our($\Pi$) obtains even better performance on all datasets and splits, which reveals the drawbacks of leveraging primarily semantic attributes for supervision and evaluation.

\subsection{Analysis and Discussion}
\noindent \textbf{Visualization of discovered Latent Prototypical Part Types.}
To verify that our parameterized $\Pi$ is able to learn diverse parts and discriminating types, we visualize some exemplar parts and types from CUB dataset in Figure~\ref{fig:cluster_example}. We observe that the examples in each type are visually similar to each other, but distinguishable to humans across different classes. When provided with the examples in each type, humans can score the existence of a type, i.e., $\pi_x(k|m)$, thereby bypassing the  proposed visual oracle. Noticeably, the part feature model $f_m(\cdot)$ is able to detect some semantic parts. For example, in Figure~\ref{fig:cluster_example}, Part-1 detects the face (or eye), Part-2 detects the beak and Part-3 tends to detect the body texture representation of birds. These semantic parts are easier to be linked to the semantic attributes, and hence our visual semantic embedding is able to close the gap between the high-dimensional visual feature and the semantic space.

It is worth noting that in Part-4, different semantic parts, like head, chin, wins and legs, are discovered in different types. We find it reasonable since the same semantic parts may not appear in different classes. The situation could be even more common in coarse-grained recognition, like a chair is not likely to have an engine. Moreover, our model tries to learn the most discriminating part via the loss $\phi_{XY}(f_m(x),y)$. The same semantic part which is the most discriminating to one class is possibly not important to another class. This phenomenon also won't cause any problem for supervision because the visual oracle is based on visual features, while for a human being this is unlikely to arise.

\noindent \textbf{Structured vs. Flat Visual Supervision} In our formulation, both the feature model $f_m(x)$ and the  mixture model $\Pi(x)$ are {\it structured}. That is, each part feature $f_m(x)$ is represented by a unique mixture model $\Pi_m$ conditioned on the part $m$. The visual oracle's supervision $\Pi_{vo}$ is also structured in a similar way. However, a different strategy is to take a flat representation and supervision: drop the part-based representation by replacing $f_m(x)$ with the global feature $E(x)$, and collapse the structured $\Pi_{vo}$ into a single list representation.
Such a flat supervision requires no part-wise features and 
its result is reported in Table \ref{tab:flat}. We observe that flattening the latent structured visual embedding as a single vector, which mirrors the common usage of semantic attributes, suffers from a slight performance drop from Ours($\Pi$) since the rich part structure information is lost. However, note that, compared to the noisy semantic supervision (see competitors in Table \ref{tab:gzsl} and \ref{tab:zsl}), the flat visual supervision still dominates competing methods.

\noindent \textbf{Visual vs. Semantic Supervision.}
To further justify the effectiveness of the proposed latent visual embedding for GZSL supervision, we took an existing state-of-the-art approach, DEVISE\cite{frome2013devise}, and re-trained it under the visual supervision $\Pi_{vo}$. As shown in Table \ref{tab:our_attr}, the proposed visual supervision also boosts DEVISE's GZSL performance, especially on AWA2, e.g. a 13.4\% absolute improvement in the harmonic mean. The result demonstrates that the proposed latent visual embedding, as a supervision type, is effective and generalizable even to non-attention methods.

\section{Conclusion}
In this paper we proposed a novel Zero-Shot learning (ZSL) method. Our method unlike many existing works neither synthesizes unseen examples nor uses any unseen semantic information during training. We claim that semantic gap exists because visual features employed in prior work is not semantic leading to significant drop in accuracy. To bridge this semantic gap we proposed a new statistical model for embedding a visual instance into a low-dimensional probability matrix. Our insight is based on the fact that analogous to how a semantic component measures the likeliness of the attribute arising in an object, so also, our mixture component conveys visual likeliness by scoring how similar a part type is relative to proto-typical part types of other instances in the training set. To further reduce semantic noise we propose a novel visual oracle for supervision in lieu of semantic supervision. We tabulate results on a number of benchmark datasets demonstrating significant improvement in accuracy over state-of-art under both semantic and visual supervision.

\section*{Acknowledgement}
The authors would like to thank the Area Chair and the reviewers for their constructive comments. This work was supported by the Office of Naval Research Grant N0014-18-1-2257, NGA-NURI HM1582-09-1-0037 and the U.S. Department of Homeland Security, Science and Technology Directorate, Office of University Programs, under Grant 2013-ST-061-ED0001.

{\small
\bibliographystyle{ieee}
\bibliography{related}
}

\end{document}